\documentclass[letterpaper, 10 pt, conference]{ieeeconf}  

\IEEEoverridecommandlockouts                              
\title{\LARGE \bf
Reinforcement Learning with Quasi-Hyperbolic Discounting}

\author{S.R. Eshwar \and Mayank Motwani \and Nibedita Roy \and Gugan Thoppe
\thanks{S.R. Eshwar, Nibedita Roy, and Gugan Thoppe are with Department of Computer Science and Automation, Indian Institute of Science, Bengaluru, India.  S.R. Eshwar was supported by the Prime Minister’s Research Fellowship (PMRF). Nibedita Roy and Gugan Thoppe were supported in part by the Walmart Centre for Tech Excellence and  the Indo-French Centre for the Promotion of Advanced Research---CEFIPRA (7102-1). Gugan Thoppe's research was also supported in part by DST-SERB's Core Research Grant CRG/2021/008330, the Kotak-IISc AI/ML Centre, and the Pratiksha Trust Young Investigator Award. 
        {\tt\small Emails: \{eshwarsr,nibeditaroy,gthoppe\}@iisc.ac.in}}%
\thanks{Mayank Motwani is with Department of Computer Science and Engineering, Indian Institute of Technology, Bombay, India. {\tt\small Email: 22b1052@iitb.ac.in}}
}
\usepackage{times}
\usepackage{bbm}
\usepackage{amsmath,amssymb}
\usepackage{graphicx, url}
\usepackage{xcolor}
\usepackage{tikz}
\usetikzlibrary{automata,arrows,positioning,calc}
\usepackage{nicematrix}
\usepackage{diagbox}
\usepackage{subcaption}
\newsavebox{\largestimage}
\usepackage{algorithm}
\usepackage{algorithmic}
\newcommand{\alglinelabel}{%
  \addtocounter{ALC@line}{-1}
  \refstepcounter{ALC@line}
  \label
}

\let\labelindent\relax 
\usepackage{enumitem}

\usepackage{amsthm}
\usepackage{mathtools} 
\usepackage[justification=centering]{caption}

\newcommand{\bE}{\mathbb{E}}

\newcommand{\bR}{\mathbb{R}}

\newcommand{\cA}{\mathcal{A}}
\newcommand{\cP}{\mathcal{P}}
\newcommand{\cR}{\mathcal{R}}
\newcommand{\cS}{\mathcal{S}}
\newcommand{\cU}{\mathcal{U}}

\newcommand{\pis}{\pi^*}

\newcommand{\hW}{\hat{W}}
\newcommand{\rx}{r_{\max}}
\newcommand{\supp}{\textnormal{supp}}

\newtheorem{theorem}{Theorem}

\newtheorem{proposition}{Proposition}

\newtheorem{remark}[theorem]{Remark}

\DeclareMathOperator*{\argmax}{arg\,max}

\begin{document}

\maketitle
\thispagestyle{empty}
\pagestyle{empty}


\begin{abstract}
Reinforcement learning has traditionally been studied with exponential discounting or the average reward setup, mainly due to their mathematical tractability. However, such frameworks fall short of accurately capturing human behavior, which has a bias towards immediate gratification. Quasi-Hyperbolic (QH) discounting is a simple alternative for modeling this bias. Unlike in traditional discounting, though, the optimal QH-policy, starting from some time $t_1,$ can be different to the one starting from $t_2.$ Hence, the future self of an agent, if it is naive or impatient, can deviate from the policy that is optimal at the start, leading to sub-optimal overall returns. To prevent this behavior, an alternative is to work with a policy anchored in a Markov Perfect Equilibrium (MPE). In this work, we propose the first model-free algorithm for finding an MPE. Using a two-timescale analysis, we show that, if our algorithm converges, then the limit must be an MPE. We also validate this claim numerically for the standard inventory system with stochastic demands. Our work significantly advances the practical application of reinforcement learning.  
\end{abstract}



\section{Introduction}
\label{sec:introduction}

\begin{figure*}
     \centering
    \savebox{\largestimage}{\includegraphics[width=0.35\textwidth]{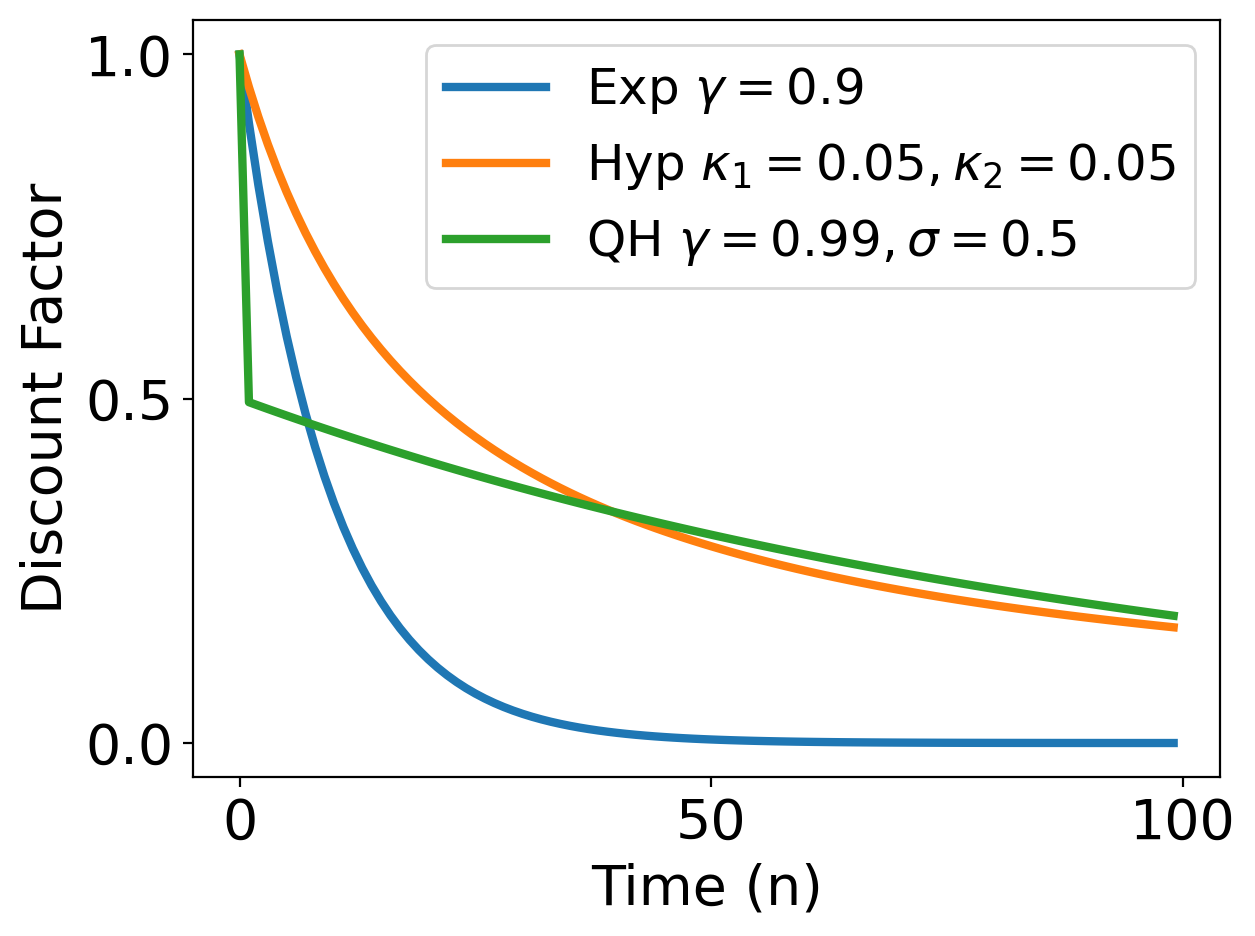}}%
     \begin{subfigure}[b]{0.35\textwidth}
         \centering
        \usebox{\largestimage}
         \caption{Comparison of discount factors.}
          \label{fig:discount_factors_comparison}
     \end{subfigure}
     \hfill
     \begin{subfigure}[b]{0.3\textwidth}
         \centering
         \raisebox{\dimexpr.5\ht\largestimage-.5\height}{
         \begin{tikzpicture}[->, >=stealth', auto, semithick, node distance=2.5cm]
	\tikzstyle{every state}=[fill=white,draw=black,thick,text=black,scale=1]
	\node[state] (1) {$1$};
        \node[state] (2) [right of=1] {$2$};  
        \path (1) edge[loop left]  node{$0.5$, $a_2$, $2$} (1)
            edge[bend left=55 ]  node{$0.5$, $a_2$, $2$}   (2)
         (2)
            edge[bend left=55] node{$1$, $a_1$, $17$}     (1);
            \draw[->] (1) to node{$1$, $a_1$, $0$}(2);

    \end{tikzpicture}
    }
         \caption{Two-state MDP example.}
         \label{fig:two-state.MDP}
     \end{subfigure}
     \hfill
     \begin{subfigure}[b]{0.3\textwidth}
         \centering
         \begin{tabular}{|p{5ex}|p{4.5ex}|p{4.5ex}|p{4.5ex}|}
        \hline
        & & & \\[-2ex]
        & \hfil $\bar{f}$ & \hfil $\bar{g}$ & \hfil $\bar{h}$ \\
        & & & \\[-2ex]
        \hline
        & & & \\[-3ex]
        \hfil $(1, a_1)$ & \hfil 18.88  & \hfil 16.83 & \hfil 18.00 \\[-1ex]
        \hfil $(1, a_2)$ & \hfil 19.00 & \hfil 16.77 & \hfil 18.00 \\[-1ex]
        \hfil $(2, a_1)$ & \hfil 32.11 & \hfil 29.56 & \hfil 31.00 \\[1ex]
        \hline 
    \end{tabular}
    \vspace{6ex}
   
         \caption{Q-values under QH-discounting.}
        \label{tab:QH.Q-Values}
     \end{subfigure}
        \caption{ \eqref{fig:discount_factors_comparison} Comparison of discount factors under exponential, hyperbolic, and quasi-hyperbolic discounting models. \eqref{fig:two-state.MDP} A two-state MDP example where the action set of state $1$ is $\{a_1, a_2\},$ while that of state $2$ is $\{a_1\}.$  For each tuple on the arrow, the first element is the probability of the transition, the second is the action taken, and the third is the instantaneous reward.  \eqref{tab:QH.Q-Values} Q-values under QH-discounting for the policies $\bar{f} \equiv (f, f, \ldots), \bar{g} \equiv (g, g, \ldots),$ and $\bar{h} \equiv (h, h, \ldots),$ where $f(1) = f(2) = g(2) = h(2) = a_1,$ while $g(1) = a_2$ and $h(a_1|1) = h(a_2|1) = 0.5.$ Each row in the table refer to the $(s, a)$ pairs, while the columns represent the corresponding policies. 
        }
   
        \label{cap:defn.f.g.h}

\end{figure*}

Reinforcement Learning (RL) \cite{sutton2018reinforcement, Bertsekas} looks at identifying a policy/strategy for an agent to optimally complete a task with sequential decisions. So far, a strategy $\bar{\pi}$'s optimality has been decided based on either the expected exponentially discounted sum or the long-term average of the sequence of rewards received under that strategy. That is, based on either $\sum_{n = 0}^\infty \gamma^n r_n$ or $\lim_{T \to \infty} \frac{1}{T} \sum_{n = 0}^{T - 1} r_n,$ where $r_n$ is the expected reward under policy $\bar{\pi}$ at time $n$ and $\gamma \in [0, 1).$ Exponential discounting is preferred when the agent has impatience, i.e., immediate gains have emphasis over future gains, with the emphasis level decided by the $\gamma$ value. In contrast, the average of the rewards is preferred when the present and future rewards are to be treated equally. However, evidence is now growing that these discounting ideas fail to  model human behaviors accurately \cite{dhami2016foundations}.

Humans are known to be impatient over shorter horizons, but not so much over longer horizons. That is, we have a bias towards instant gratification. This can be understood from the famous example by Richard Thaler \cite{thaler1981some}, who said, ``Most people would prefer one apple today to two apples tomorrow, but they prefer two apples in 51 days to one in 50 days." Observe that there is a reversal of preferences when the time frame shifts. This phenomenon is known as the \emph{common difference effect} \cite{dhami2016foundations}. Such preference reversals cannot happen under a policy that is optimal with respect to either of the two traditional discounting models. This is because of their time-consistent nature \cite{sutton2018reinforcement}, i.e, this optimal policy remains optimal even when reconsidered from some later time as well. This demonstrates the limitations of these discounting models in explaining human behaviors.

Hyperbolic discounting  \cite{loewenstein1992anomalies} is a leading candidate \cite{ainslie1975specious, cropper1992rates, frederick2002time} for explaining the common difference effect. The value of a strategy $\bar{\pi}$ under this discounting model is  $\sum_{n = 0}^\infty b_n r_n,$ where $r_n$ is defined as before and $b_n = \left(1 + \kappa_1 n\right)^{-\kappa_2/\kappa_1}$ for some $\kappa_1, \kappa_2 > 0.$ However, this form of discounting is quite complicated, making its study hard. This brings forth Quasi-Hyperbolic (QH) discounting \cite{phelps1968second, Laib}, which is a simpler and more tractable alternative. In QH discounting, $b_0 = 1$ and $b_n = \sigma \gamma^n,\  n \geq 1,$ for some $\sigma \in [0, 1]$ and $\gamma \in [0, 1).$ The symbol $\sigma$ is the short-term discount factor, while $\gamma$ is the long-term discount factor. Clearly, for $\sigma = 1,$ QH discounting matches exponential discounting. A comparison of the discount factors under exponential, hyperbolic, and quasi-hyperbolic discounting is given in Fig.~\ref{fig:discount_factors_comparison}. Unlike exponential discounting, note that there is a sharp decrease in hyperbolic and QH discount factors initially, after which they decrease more gradually. In this work, we initiate the study of RL with QH discounting. 

Under exponential discounting,  the optimal policy $\bar{\pi}_*$ is deterministic and stationary, and has a greedy relationship with its $Q$-value function \cite{sutton2018reinforcement}. However, such properties do not always hold under QH-discounting \cite{Nowak}. This can lead to a complicated agent behavior, as we now illustrate.

Consider the two-state Markov Decision Process (MDP) setup of Fig.~\ref{fig:two-state.MDP}, which is taken from \cite{Nowak}. Clearly, there are only two deterministic stationary policies here: $\bar{f} \equiv (f, f, \cdots)$ and $\bar{g} \equiv (g, g, \ldots),$ where $f$ maps state $1$ to action $a_1$, and $g$  maps state $1$ to action $a_2$, and both map state $2$ to action $a_1$. For $\sigma = 0.5$ and $\gamma = 0.8$, their $Q$-value functions under QH-discounting are given in Table~\ref{tab:QH.Q-Values}. For a policy $\bar{\pi},$ its QH Q-value function, denoted by $Q^{\sigma, \gamma}_{\bar{\pi}},$  is defined in the same way as in the exponential discounting case, but with discount factors $1, \gamma, \gamma^2, \ldots$ replaced by $1, \sigma \gamma, \sigma \gamma^2, \ldots$ In Table~\ref{tab:QH.Q-Values}, notice that $\bar{f}$ yields the highest returns from state $2.$ However, at state $1,$ 
neither $\bar{f}$ nor $\bar{g}$ shares a greedy relationship with its QH $Q$-value function. This fact implies that $g\bar{f}$ (acting as $g$ at $n=0$ and $f$ for $n \geq 1$) is the policy that yields the highest returns, starting from state $1.$ 

The optimal policies in the above example have three interesting dissimilarities compared to their counterparts in RL with exponential discounting or simple averaging. Firstly, the optimal policy varies depending on the initial state of the process. Secondly, $g \bar{f}$ is non-stationary. Thirdly, and significantly, both $\bar{f}$ and $g \bar{f}$ display time inconsistency. To elaborate the last point further, note that both $g \bar{f}$ and $\bar{f}$ advocate following $f$ at any $n \geq 1$. Now suppose, at time $n = 1,$ the MDP is in state 1 and the agent re-evaluates the optimal policy from that time onward. Then, from Table~\ref{tab:QH.Q-Values}, the agent would again discover $g \bar{f}$ to be optimal, i.e., act as per $g$ at $n = 1$ and revert to $f$ thereafter. This behavior contradicts the one that is optimal from $n = 0,$ highlighting the time inconsistency.

We now describe a complex agent behavior for the above setup, stemming from the time-inconsistent nature of the optimal policies. Consider the agent as a sequence of selves, each corresponding to a different time step $n.$ Suppose each future self is naive, i.e., unaware of the time inconsistency in the optimal policy. Alternatively, suppose they all have self-control issues and a possibility to act contrary to their own interests. In both scenarios, the following scenario could unfold. Each time the MDP visits state $1,$ the naive selves recalculate the optimal strategy from that point on and decide to act as per $g$ at that time step. Similarly, each self with control issues could also decide to act as per $g$ because (i.) it is aware that if the subsequent selves act as per $f,$ then it would receive higher returns, and (ii.) it presumes that the subsequent selves will act as per $f.$ However, if all selves act as per $g$ for all $n \geq 0,$ then Table~\ref{tab:QH.Q-Values} shows that the expected overall returns would be substantially lower (only $16.77$). 

To safeguard against the above pitfall, it is desirable to have a stationary (possibly stochastic) policy $\bar{\pi} \equiv (\pi,  \pi, \ldots)$ from which there is no incentive for deviation. 
For such a policy $\bar{\pi}$, it would then be true that
\begin{equation}
\label{e:MPE}
    Q^{\sigma, \gamma}_{\bar{\pi}}(s, \pi) = \sup_{\nu: \cS \to \Delta(\cA)}  Q^{\sigma, \gamma}_{\bar{\pi}}(s, \nu), \qquad s \in \cS,
\end{equation}
where $\cS$ (resp. $\cA$) is the MDP state space (resp. action space), $\Delta(\cA)$ is the set of distributions on $\cA,$  and $Q^{\sigma, \gamma}_{\bar{\pi}}(s, \nu) = \sum_{a \in \cA(s)} \nu(a|s)  Q^{\sigma, \gamma}_{\bar{\pi}}(s, a)$ is the average of $\bar{\pi}$'s QH Q-values for a starting state distribution $\nu: \cS \to \Delta(\cA).$ Any stationary policy $\bar{\pi}^*$ which satisfies \eqref{e:MPE} is called a Markov Perfect Equilibrium (MPE) \cite{Nowak}. For our two-state MDP example, Table \ref{tab:QH.Q-Values} shows that the stationary policy $\bar{h}$ is an MPE. For a general MDP, MPEs neither need to exist nor be unique; so far, they have been found only using analytical techniques, and that too only for simple MDPs \cite{Nowak}. 

Our goal in this work is to develop a model-free RL algorithm to identify an MPE in a finite state and action MDP. An MPE's existence here is guaranteed by the conditions outlined in \cite{Nowak}. However, finding it poses significant challenges, unlike finding an optimal policy in classical RL. Firstly, no known Bellman-type contraction mapping exists for which an MPE's Q-value function is a fixed point. Consequently, the traditional fixed-point-type methods cannot be used to find this value function. Moreover, even if this function were identified somehow, determining the MPE itself remains challenging as it lacks a straightforward relationship with its value function. Recall that the optimal policy in classical RL is greedy with respect to its Q-value function. Secondly, MPEs often are stochastic. This means the search space for an MPE encompasses all stochastic policies, which is infinitely large even for finite state and action MDPs. In contrast, for traditional discounting, the optimal policy search is confined to deterministic policies, which is a finite set (albeit growing combinatorially).

Our main contributions are as follows. We propose the first model-free RL algorithm for finding an MPE. This algorithm is a two-timescale stochastic approximation and is inspired by the recently proposed critic-actor method for classical RL \cite{bhatnagar2023actor}. Unlike the actor update in \cite{bhatnagar2023actor}, which follows a stochastic estimate of the value function's gradient, our method updates along the QH advantage function\footnote{The QH advantage function is the difference between the QH Q-value function and the state-value function. In RL with exponential discounting, the advantage function and the value function's gradient are aligned, but it is not so under QH discounting; see \eqref{th:PG-QH}.}, enabling it to find an MPE. Secondly, by building upon \cite{ramaswamy2016stochastic}, \cite{gopalan2022demystifying}, \cite{bhatnagar2023actor}, and \cite{yaji2020stochastic}, we show that if the critic-actor iterates of our algorithm converge to an isolated point $(W, \pi)$, then $\pi$ must be an MPE and $W$ its Q-value function. Finally, we provide numerical experiments in an inventory control setup, demonstrating our algorithm's success in identifying various MPEs.

\section{Setup, Goal, Algorithm, and Main Results}
In this section, we describe our problem setup, our goal, and our key contributions: the first algorithm for finding an MPE and results that describe its asymptotic convergence.

\subsection{Setup and Goal}
Let $\Delta(\cU)$ denote the set of distributions over a set $\cU.$ Our setup consists of an MDP $M \equiv (\cS, \cA, \cP, \cR, \sigma, \gamma),$ where $\cS$ and $\cA$ are finite state and finite action spaces, respectively, $\cP: \cS \times \cA \to \Delta(\cS)$ is the transition matrix, and $r: \cS \times \cA \to \bR$ is the instantaneous reward function. Further, $\sigma, \gamma \in [0, 1)$ are the parameters of QH discounting. Within the above setup, our goal is to find an MPE, i.e., a stationary policy $\bar{\pi} \equiv (\pi, \pi, \ldots)$ (henceforth denoted only by $\pi$) that satisfies the MPE relation given in \eqref{e:MPE}. 

\subsection{MPE-learning Algorithm}
\label{subsec:proposed_algo}
For a stationary policy $\pi: \cS \to \Delta(\cA),$ let  $Q_\pi^{\sigma,\gamma}$ denote the policy $\pi$'s QH Q-value function. Formally, let
\begin{multline}
\label{eq:QH_Q-value_fn}
    Q_\pi^{\sigma,\gamma}(s, a) := r(s, a) \\ 
    +  \bE \left[\sum_{n = 1}^\infty \sigma \gamma^n r(s_n, a_n) \bigg|{\genfrac{}{}{0pt}{}{s_0 = s,}{a_0 = a}}
    \right],    
\end{multline}
where $s_{n + 1} \sim \cP(\cdot|s_n, a_n)$ and $a_{n + 1} \sim \pi(\cdot|s_{n + 1}).$ Also, for $\theta \in \bR^{|\cS| \cdot |\cA|}$ and $s \in \cS,$ let $\pi_\theta(\cdot|s) = \textnormal{softmax}(\theta(s, \cdot)).$

Our novel approach for finding an MPE is given in Algorithm \ref{alg:sync-algo}. The symbol $\theta_n$ parameterizes the policy representing our MPE estimate at time $n \geq 0,$ while $W_n$ is this policy's Q-value function estimate. Hence, we refer to the $\theta_n$ update (Step~\ref{alg.eq:theta_sync_update_rule}) as the actor update, and to the $W_n$ update (Step~\ref{alg.eq:W_sync_update_rule}) as the critic update. Throughout this work, we focus on the scenario where the stepsizes $\alpha_n$ and $\beta_n,$ used in the critic and actor updates, respectively, satisfy the relation $\lim_{n \to \infty} \alpha_n/\beta_n = 0.$ This ensures the critic updates are on a slower timescale compared to the actor. Consequently, Algorithm~\ref{alg:sync-algo} falls under the category of critic-actor algorithms \cite{bhatnagar2023actor} (instead of actor-critic). We give a principled motivation for our algorithm in Section~\ref{sec:app.algo_design}.

\begin{algorithm}[tb]
   \caption{Synchronous MPE-learning}
   \label{alg:sync-algo}
\begin{algorithmic}[1]
   \STATE {\bfseries Input:} Stepsizes $(\alpha_n), (\beta_n),$ and discount factors $\sigma, \gamma$
   \STATE {\bfseries Initialize:}           $\theta_0 , W_0 \in \mathbb{R}^{|\cS| \times |\cA|}$
    \FOR{$n=0,1,2,...$} 
    \FOR{$(s,a) \in \cS \times \cA$} \alglinelabel{alg.eq:start.step}
    \STATE Observe $s' \sim \cP(\cdot|s,a)$\\[1ex]
    \STATE Sample $a' \sim \pi_{\theta_n}(\cdot|s')$\\[1ex]
    \STATE $r'_n(s,a) \leftarrow r(s',a'),\quad  W'_n(s,a) \leftarrow W_n(s',a')$ \alglinelabel{alg.eq:rn'_Wn'}\\[1ex]
    \STATE $\hat{W}_n^{\theta_n}(s,a) \leftarrow \langle \pi_{\theta_n}(\cdot|s), \ W_n(s, \cdot) \rangle$\\[1ex]
   \ENDFOR \alglinelabel{alg:eq:final.step} \\[1ex]
   \STATE \alglinelabel{alg.eq:W_sync_update_rule} $W_{n+1} \leftarrow W_n + \alpha_n \left[ r - (1-\sigma) \gamma r'_n + \gamma W_n'- W_n\right]$\\[1ex]
   \STATE \alglinelabel{alg.eq:theta_sync_update_rule} $\theta_{n+1} \leftarrow \theta_n + \beta_n \left[ W_n - \hW_n^{\theta_n} \right]$
   \ENDFOR
\end{algorithmic}
\hrulefill \\
\textit{Steps~\ref{alg.eq:W_sync_update_rule} and \ref{alg.eq:theta_sync_update_rule} define the algorithm's update rules, while Steps~\ref{alg.eq:start.step} to \ref{alg:eq:final.step} setup the necessary vectors for these updates. The angle bracket $\langle \cdot, \cdot \rangle$ denotes inner product.}
\end{algorithm}

\subsection{Main  Results}
\label{subsec:main_conj_other_results}
We first state our assumptions. 
{
\renewcommand{\theenumi}{$\cA_\arabic{enumi}$}
\begin{enumerate}[leftmargin=*]
\item \label{a:stepsize} \textbf{Stepsizes: }$(\alpha_n)_{n \geq 0}$ and $(\beta_n)_{n\geq0}$ are two sequences of monotonically decreasing positive real numbers such that 
        (i) $\alpha_0 \leq 1,\beta_0 \leq 1;$ 
        (ii) $\sum_{n=0}^{\infty} \alpha_n=\sum_{n=0}^{\infty} \beta_n=\infty,$ but $\sum_{n=0}^{\infty} (\alpha_n^2+ \beta_n^2) <\infty;$ and (iii) $\lim_{n\rightarrow \infty} (\alpha_n/\beta_n) = 0.$ 

\item \label{a:reward} \textbf{Bounded reward:} There exists $\rx > 0$ such that $|r(s,a)| \leq \rx$ for all $s \in \cS$ and $a \in \cA.$ 
\end{enumerate}
}

Next, we define a set-valued map. For $W \in \bR^{|\cS||\cA|},$ let $\lambda(W)$ be the (convex) set of stochastic policies given by 
\begin{multline}
\label{e:lambda.defn}
    \lambda(W) := \bigg\{\pi: \cS \to \Delta (\cA): \sum_{a \in \cA} \pi(a|s) = 1 \text{ and }  \\
    \supp(\pi(\cdot|s)) \subseteq \argmax W(s, \cdot)\}\ \forall s \in \cS \bigg\}.
\end{multline}

Our main result can now be stated as follows. Let $\|\cdot\|$ be the standard $\ell_\infty$ norm.

\begin{theorem}
\label{th:main_result}
Suppose \ref{a:stepsize} and \ref{a:reward} are true. Then the following statements hold for the iterates $(W_n)$ and $(\theta_n)$ obtained from Algorithm~\ref{alg:sync-algo}: 
\begin{enumerate}[label=(\roman*)]
    \item \label{st:W.stability} $(W_n)$ is stable, i.e., $\sup_n \|W_n\| < \infty$\ a.s.;
    
    \item \label{st:fast.iterate.convergence} 
    If $(W_n)$ held constant at $W,$ then $\pi_{\theta_n} \to \lambda(W);$ and
    
    \item \label{st:slow.iterate.convergence} If $(W_n, \pi_{\theta_n}) \rightarrow (W^*,\pi^*)$, then $\pi^*$ must be an MPE and $W^*$ must be the QH $Q$-value function of $\pi^*$. 
\end{enumerate}
\end{theorem}

\begin{remark}
    Our first statement shows that our $(W_n)$ iterates would be uniformly bounded. Next, for the case where the $(W_n)$ sequence is (hypothetically) held constant at $W,$ our second statement shows that the actor updates would lead to a policy that is greedy with respect to $W.$ This result is important since the two-timescale nature of the algorithm, viz. the condition $\alpha_n/\beta_n \to 0,$ implies the $(W_n)$ iterates are quasi-static (or slowly changing) from the viewpoint of the $\theta_n$ updates. Our final statement shows that if our algorithm converges, then the limit must correspond to an MPE.
\end{remark}

\begin{figure*}[ht]
     \centering
     \begin{subfigure}[b]{0.3\textwidth}
         \centering
         \includegraphics[width=\textwidth]{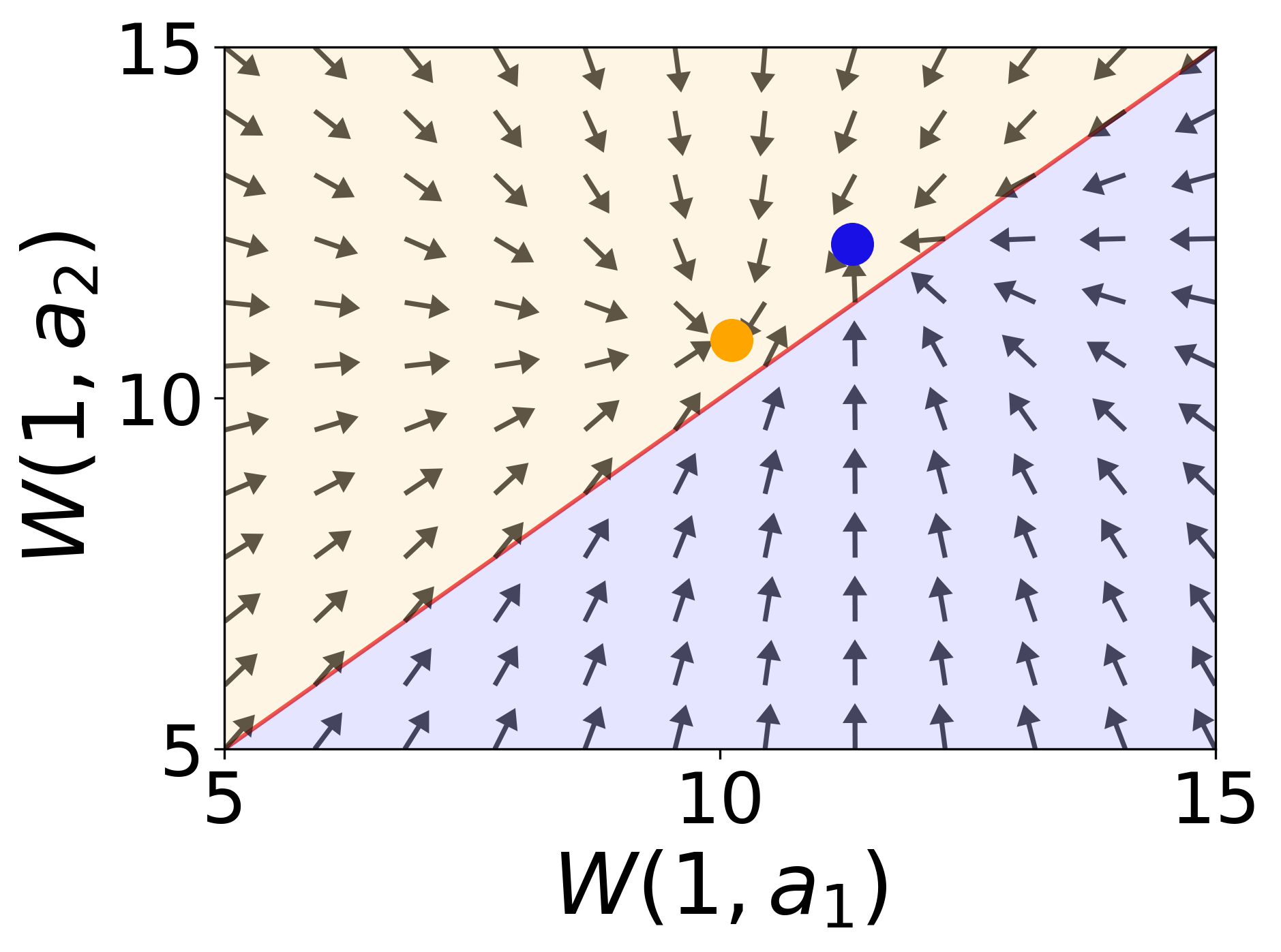}
         \caption{$\sigma=0.3$}
         \label{fig:sigma_0.3}
     \end{subfigure}
     \hfill
     \begin{subfigure}[b]{0.3\textwidth}
         \centering
         \includegraphics[width=\textwidth]{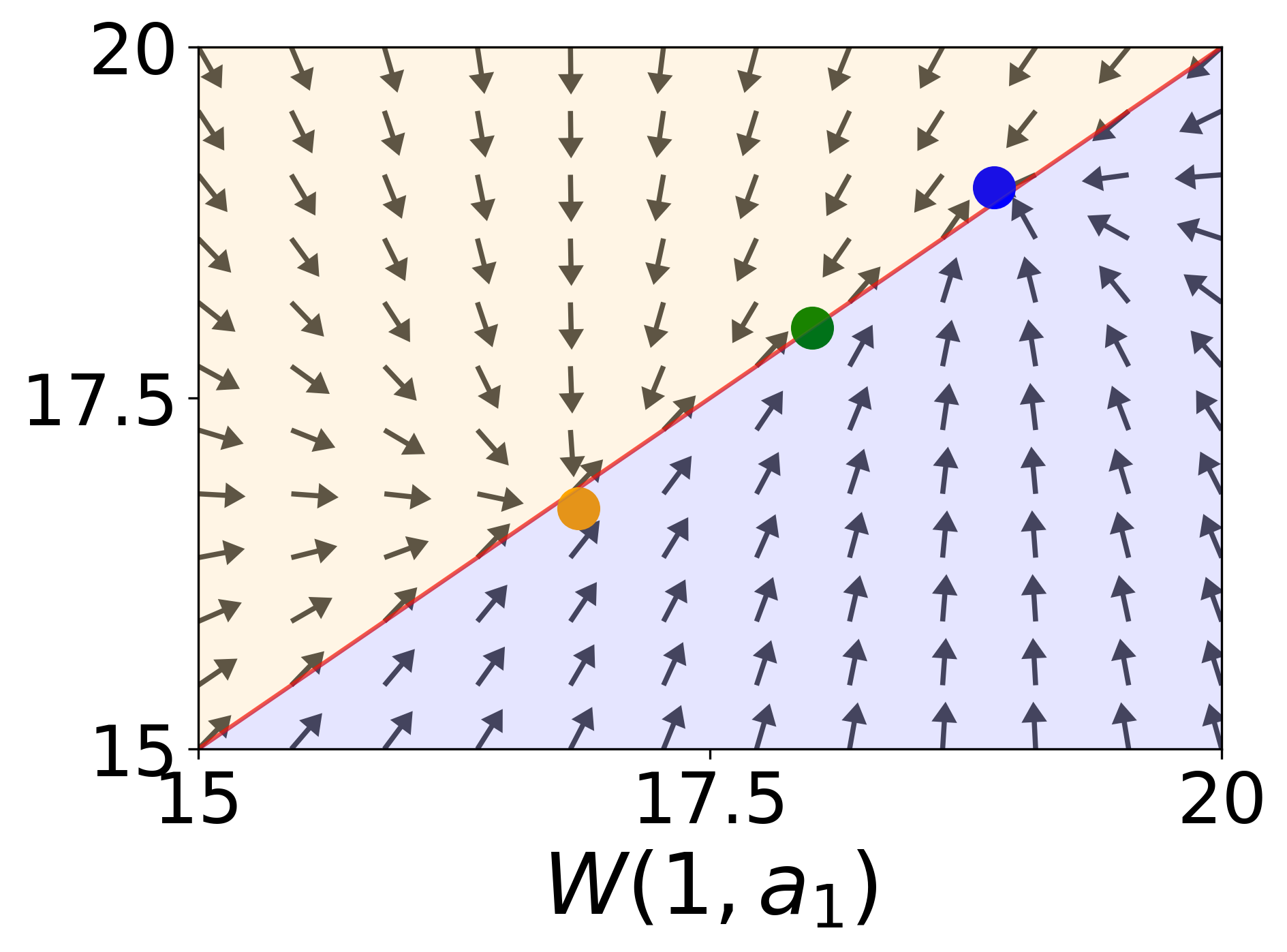}
         \caption{$\sigma=0.5$}
         \label{fig:sigma_0.5}
     \end{subfigure}
     \hfill
     \begin{subfigure}[b]{0.3\textwidth}
         \centering
         \includegraphics[width=\textwidth]{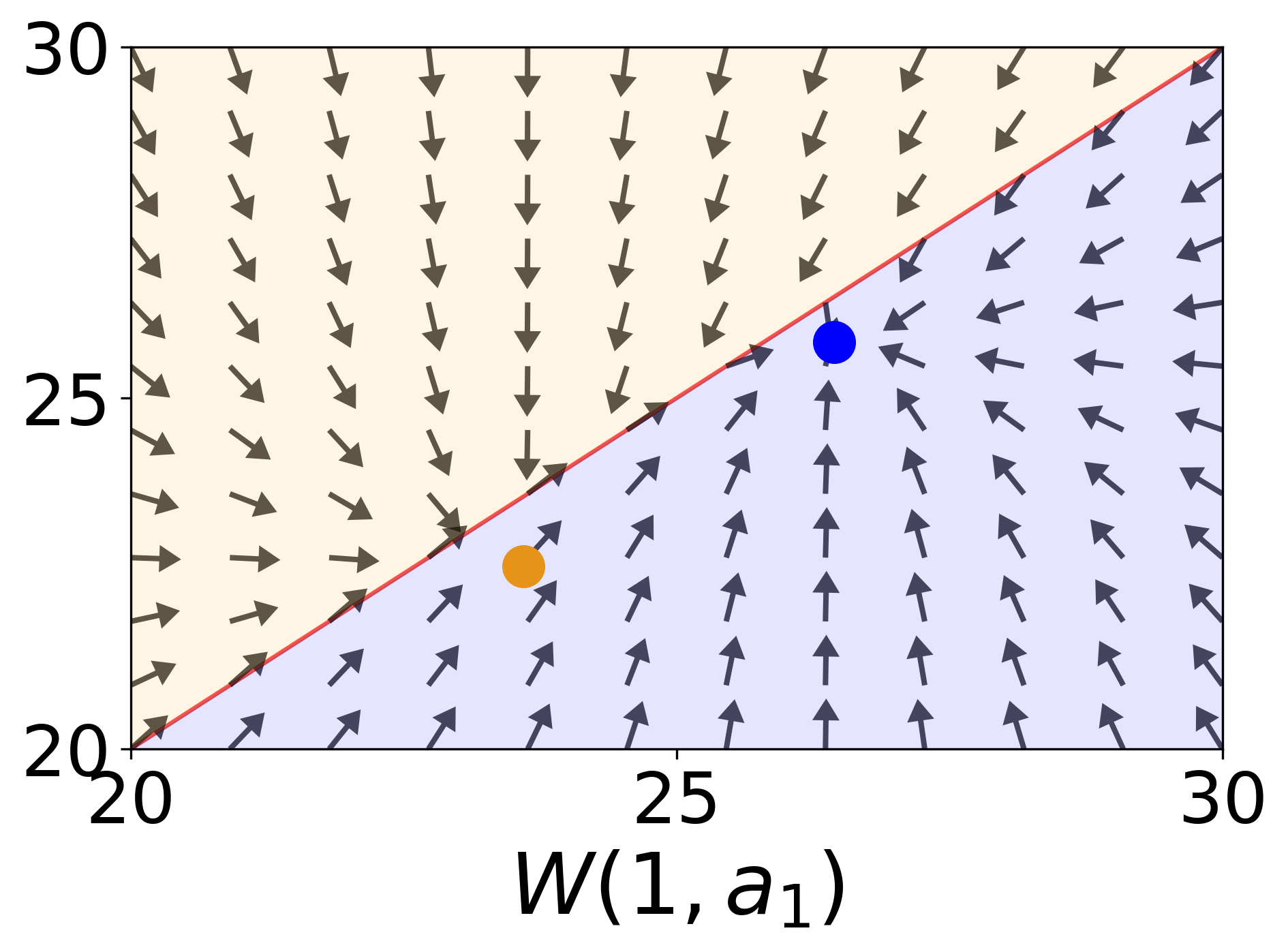}
         \caption{$\sigma=0.7$}
         \label{fig:sigma_0.7}
     \end{subfigure}
        \caption{Vector fields for the DI in \eqref{eq:slow_ts_DI} for the MDP given in Fig.~\ref{fig:two-state.MDP} with $\gamma = 0.8$ and $\sigma$ values of $0.3, 0.5,$ and $0.7.$  Here, the orange dot is the vector $Q^{\sigma,\gamma}_{\bar{g}},$ blue is $Q^{\sigma,\gamma}_{\bar{f}},$ while green is $Q^{\sigma,\gamma}_{\bar{h}},$ where $\bar{f}, \bar{g}$ and $\bar{h}$ are as in Fig.~\ref{cap:defn.f.g.h}'s caption. }
        \label{fig:vector field}
\end{figure*}

\section{Our Algorithm Design}
\label{sec:app.algo_design}
We now motivate the critic (Step~\ref{alg.eq:W_sync_update_rule}) and actor (Step~\ref{alg.eq:theta_sync_update_rule}) update rules of our proposed algorithm and explain how they enable MPE estimation. 

Our critic update step is based on the temporal difference idea for minimizing the QH Bellman error at time $n$, i.e., the critic tries to find a $W$ that minimizes the gap between $T^{\pi_{\theta_n}} (W)$ and $W$ for the given $\theta_n.$ Hence, $W_n$ can be viewed as an estimate of the QH Q-value function of $\pi_{\theta_n}.$

Our actor or the $\theta_n$ update is approximately along the QH advantage function  $A^{\sigma, \gamma}_{\pi_{\theta_n}}$ of $\pi_{\theta_n},$ where $A^{\sigma, \gamma}_{\pi}(s, a) = Q^{\sigma, \gamma}_{\pi}(s, a) - \langle \pi(\cdot|s), Q^{\sigma, \gamma}_{\pi}(s, \cdot)\rangle$ for any $\pi, s,$ and $a.$ We say approximately because $\theta_n(s, a)$ is actually updated along $W_n(s, a) - \langle \pi_{\theta_n}(\cdot|s), W_n(s, \cdot) \rangle$ and $W_n$ is only an estimate of $Q^{\sigma, \gamma}_{\pi_{\theta_n}}.$ Our main motivation to use the advantage function for updating $\theta_n$ is to ensure that  $\pi_{\theta_n}$ is asymptotically greedy with respect to $W_n;$ see Theorem~\ref{th:main_result}.\ref{st:fast.iterate.convergence}. Even in RL with exponential discounting, the actor updates are along the corresponding advantage function estimate of the current policy \cite{sutton1999policy}. However, the advantage function  there aligns with the gradient of the state value function and enables discovery  of the optimal policy. In QH discounting, though, this alignment does not hold, as we show next. 

Let $A^{\gamma}_{\pi}(s,a) = Q^{\gamma}_{\pi}(s,a) - \langle \pi(\cdot|s), \ Q^{\gamma}_{\pi}(s,\cdot) \rangle$ be the policy $\pi$'s advantage function under exponential discounting with $\gamma$  discount factor. Similarly, let $A^{0}_{\pi}(s,a)$ be the analogous $\gamma = 0$ expression. Now, for $\theta \in \bR^{|\cS|\cdot |\cA|},$ if $\eta^{\sigma,\gamma}_{\pi_\theta}(\mu):= \mathbb{E}_{s\sim \mu, a \sim \pi_{\theta}}[Q^{\sigma,\gamma}_{\pi_\theta}(s,a)]$ denotes an averaging of $\pi_\theta$'s QH Q-value function with respect to some fixed initial state distribution $\mu,$ then we have that
\begin{equation}
\label{th:PG-QH}
    \begin{split}
        \frac{\partial \eta_{\pi_\theta}^{\sigma,\gamma}(\mu)}{\partial \theta(s,a)} = {} & (1-\sigma) \mu(s) \pi_\theta(a|s) A^0_{\pi_\theta}(s,a) \\
        & \quad + \frac{\sigma}{1-\gamma} d_\mu^{\pi_\theta}(s) \pi_\theta(a|s) A^\gamma_{\pi_\theta}(s,a).
    \end{split}
\end{equation}
Clearly, the RHS does not equal 
\[
    A^{\sigma, \gamma}_{\pi_\theta}(s, a) \equiv (1 - \sigma) A^0_{\pi_\theta}(s, a) + \sigma A^{\gamma}_{\pi_\theta}(s, a),
\]
which itself holds since 
\begin{align*}
    Q^{\sigma, \gamma}_{\pi_\theta} &(s, a)\\
    %
    %
    = {} & \bE \left[(1 - \sigma) r(s_0, a_0) + \sum_{n = 0}^\infty \sigma \gamma^n r(s_n, a_n) \bigg|{\genfrac{}{}{0pt}{}{s_0 = s,}{a_0 = a}}
    \right].
\end{align*}
This verifies our aforementioned claim that the advantage function does not align with the gradient. In fact, this also explains why our algorithm does not track the optimal policy, unlike the classical critic-actor method.  

Finally, we provide evidence for why our algorithm's convergence implies that the limit must be an MPE. Let $T: \bR^d \to 2^{\bR^d}$ be given by
\begin{equation}
\label{e:T.defn}
    T(W) := \{T^\pi(W): \pi \in \lambda(W)\},
\end{equation}
where $\lambda$ is as in \eqref{e:lambda.defn} and $T^\pi: \bR^{|\cS||\cA|} \to \bR^{|\cS||\cA|}$ is the QH Bellman operator for the policy $\pi.$ That is, 
\begin{multline}
\label{e:T.mu.defn}
    T^\pi(W) (s,a) =   r(s,a)\  + 
    \gamma \sum_{s',a'} \cP(s'|s,a) \pi(a'|s') \\
    \times \left[ -(1-\sigma) r(s',a') + W(s',a') \right].
\end{multline}
Then, for the case where our algorithm's iterates converge, it can be shown (\cite{bhatnagar2023actor, borkar2009stochastic, yaji2020stochastic}, \cite{gopalan2022demystifying}) by building upon Theorem~\ref{th:main_result}.\ref{st:fast.iterate.convergence} and the two-timescale nature of our algorithm that our $(W_n)$ iterates asymptotically track the solution trajectories of the Differential Inclusion (DI) \cite{aubin2012differential}
\begin{equation}
\label{eq:slow_ts_DI}
    \dot{W}(t) \in T (W(t)) - W(t).
\end{equation}
The limiting set-valued dynamics arises because the same $W$ can have multiple greedy policies. 

Tracking the above DI is valuable for MPE estimation, and we demonstrate this in two ways. On the one hand, we present the following general result, showing that every MPE's QH Q-value function is a zero of this limiting DI. 
    
\begin{proposition}
\label{prop:MPE_is_0_of_DI}
    For any MPE $\pi^*,$ we have $\pis \in \lambda (Q^{\sigma,\gamma}_{\pi^*})$ which, in turn, implies that $0 \in T(Q^{\sigma,\gamma}_{\pi^*}) - Q^{\sigma,\gamma}_{\pi^*}.$
\end{proposition}
On the other hand, we discuss this DI's vector field plot in Fig.~\ref{fig:vector field} for the specific two-state MDP given in Fig.~\ref{fig:two-state.MDP}, highlighting the complexities of MPE estimation and how this DI addresses them. The three subplots in Fig.~\ref{fig:vector field} correspond to different values of $\sigma:$ $0.3, 0.5,$ and $0.7,$ respectively, while $\gamma = 0.8$ in all. In all these subplots, there is a blue and an orange colored region. These are the greedy regions associated with the policies $\bar{f}$ and $\bar{g},$ respectively (see Fig.~\ref{cap:defn.f.g.h}'s caption for $\bar{f}$ and $\bar{g}$'s  definition). That is, for any vector $W$ in the blue region $W(1, f(1)) = W(1, a_1) \geq W(1, a_2) = W(1, g(1))$ and the reverse holds for any vector in the orange region. Hence, for any $W$ in the interior of the blue (resp. orange) region, $\lambda(W)$ consists\footnote{We mean $\lambda(W)$ contains the stochastic representation of $f.$} of only $f$ (resp. $g$) and the RHS of \eqref{eq:slow_ts_DI} is $T^f (W) - W$ (resp. $T^g(W) - W$),  where $T^f$ and $T^g$ are defined as in \eqref{e:T.mu.defn}. Because the greedy policy is different in the two colored regions, the local dynamics discontinuously changes across the $x = y$ boundary line. As pointed out before, the DI in \eqref{eq:slow_ts_DI} handles this discontinuity by allowing both the update directions (and also their convex combinations) at the boundary.

In Fig.~\ref{fig:vector field}(a),  any solution trajectory starting in the blue region is driven towards the blue dot, which represents $Q_{\bar{f}}^{\sigma, \gamma}.$ This is not surprising since we use QH temporal difference learning for updating $(W_n).$ However, once the trajectory crosses over to the orange region, the driving function changes and the trajectory is now driven towards the orange dot, which represents $Q_{\bar{f}}^{\sigma, \gamma}.$ For $\sigma = 0.3,$ it can be shown that $\bar{g}$ is the only MPE and tracking the solution trajectory of our DI helps in finding this MPE's QH Q-value. Fig.~\ref{fig:vector field}(c) can be interpreted similarly. In Fig.~\ref{fig:vector field}(b), i.e., for the case $\sigma = 0.5,$ it can be shown that neither $\bar{g}$ nor $\bar{f}$ is an MPE. Instead, the stochastic policy $\bar{h}$ (see Fig.~\ref{cap:defn.f.g.h}'s caption) is an MPE and its Q-value sits on the boundary. In this case, the driving function in either region pushes the solution trajectory towards the other and, hence, eventually, it is forced to converge to the green dot, which is $Q_{\bar{h}}^{\sigma, \gamma}.$ Thus, tracking the solution trajectories of our DI again helps.

\begin{table*}[!ht]
\centering
\caption{\textbf{MPEs and other Policies}. $\boldsymbol{\pi^*_1}$, $\boldsymbol{\pi^*_2}$, $\boldsymbol{\pi^*_3}$  correspond to three different MPEs, whereas $\boldsymbol{\pi^*}$ is the optimal policy (obtained via the gradient-based critic-actor method) and $\boldsymbol{\pi_\text{n}}$ is the associated sub-optimal policy that a naive manager will follow; the $(s, a)$-th entry in each matrix is probability of picking action $a \in \{0, 1, 2\}$ at state $s \in \{0, 1, 2\}.$ }
\label{tab:policy_matrices}
\begin{equation*}
\centering
\boldsymbol{\pi^*_1 }= \begin{bmatrix}
0.00 & 0.53 & 0.47 \\
0.53 & 0.47 & - \\
1.00 & - & -\\
\end{bmatrix}, 
\boldsymbol{\pi^*_2} = \begin{bmatrix}
0.0 & 0.8 & 0.2 \\
0.0 & 1.0 & - \\
1.0 & - & - 
\end{bmatrix},
\boldsymbol{\pi^*_3 }= \begin{bmatrix}
0.0 & 0.3 & 0.7 \\
1.0 & 0.0 & - \\
1.0 & - & -\\
\end{bmatrix},
\boldsymbol{\pi^*} = \begin{bmatrix}
0.0 & 0.0 & 1.0 \\
0.0 & 1.0 & - \\
1.0 & - & -\\
\end{bmatrix},
\boldsymbol{\pi_\text{n}}= \begin{bmatrix}
0.0 & 1.0 & 0.0 \\
1.0 & 0.0 & - \\
1.0 & - & -\\
\end{bmatrix}
\end{equation*}
\end{table*}

\begin{table*}[!ht]
\caption{\textbf{QH Q-value functions}. The numbers in bold represent the \( Q^{\sigma,\gamma} \)-values for actions recommended by the respective policies. In MPEs, no other actions have higher values than those in bold, so the agent has no incentive to deviate from them. In contrast, under $\pi^*$ from the Vanilla QH Policy Gradient Algorithm, the underlined actions have higher values, hence an agent may deviate from $\pi^*$.}
\begin{minipage}{.3\linewidth}
    \centering
    \caption*{$Q^{\sigma,\gamma}_{\pi^*_{1}}$ values}
\begin{tabular}{ |c|c|c|c| } 
 \hline
 \backslashbox{$s$}{$a$} & \hfil0 & \hfil1 & \hfil2 \\
 \hline
 \hfil0 & \hfil897.5 &  \hfil\textbf{1053} &  \hfil\textbf{1053} \\
 \hfil1  & \hfil\textbf{1553} &  \hfil\textbf{1553} & \hfil-\\
 \hfil2 & \hfil\textbf{2053} & \hfil- & \hfil- \\
 \hline
\end{tabular}
\label{tab:IC.Q.values}
\vspace{1.5em}
\end{minipage}
\hfil
\begin{minipage}{.3\linewidth}
\centering
\caption*{$Q^{\sigma,\gamma}_{\pi^*_{2}}$ values}
\begin{tabular}{|c|c|c|c| } 
 \hline
 \backslashbox{$s$}{$a$} & \hfil0 & \hfil1 & \hfil2 \\
 \hline
 \hfil0 & \hfil873.75 &  \hfil\textbf{1040.5} &  \hfil\textbf{1040.5} \\
 \hfil1  & \hfil1540.5 &  \hfil\textbf{1540.5} & -\\
 \hfil2 & \hfil\textbf{2040.5} & - & - \\
 \hline
\end{tabular}
\vspace{1.5em}
\end{minipage}
\hfil
\begin{minipage}{.3\linewidth}
\centering
\caption*{$Q^{\sigma,\gamma}_{\pi^*_{3}}$ values}
\begin{tabular}{|c|c|c|c|} 
 \hline
 \backslashbox{$s$}{$a$} & \hfil0 & \hfil1 & \hfil2 \\
 \hline
 \hfil0 & \hfil918.64 &  \hfil\textbf{1064.125} &  \hfil\textbf{1064.125} \\
 \hfil1  & \hfil\textbf{1564.125} &  \hfil1564.125 & -\\
 \hfil2 & \hfil\textbf{2064.125} & - & - \\
 \hline
\end{tabular}
\vspace{1.5em}
\end{minipage}

\hfil
\begin{minipage}{0.3\linewidth}
    \centering
    \caption*{$Q^{\sigma,\gamma}_{\pi^*}$ values}
\centering
\begin{tabular}{ |c|c|c|c| } 
 \hline
 \backslashbox{$s$}{$a$} & \hfil0 & \hfil1 & \hfil2 \\
 \hline
 \hfil0 & \hfil1080 &  \hfil\underline{1235.5} &  \hfil\textbf{1228} \\
 \hfil1  & \hfil\underline{1735.5} &  \hfil\textbf{1728} & \hfil-\\
 \hfil2 & \hfil\textbf{2228} & \hfil- & \hfil- \\
 \hline
\end{tabular}
\end{minipage}
\hfil
\begin{minipage}{0.3\linewidth}
    \centering
\caption*{$Q^{\sigma,\gamma}_{\pi_{\text{n}}}$ values}
\centering
\begin{tabular}{ |c|c|c|c| } 
 \hline
 \backslashbox{$s$}{$a$} & \hfil0 & \hfil1 & \hfil2 \\
 \hline
 \hfil0 & \hfil675 &  \hfil\textbf{830.5} &  \hfil839.6 \\
 \hfil1  & \hfil\textbf{1330.5} &  \hfil1339.6 & \hfil-\\
 \hfil2 & \hfil\textbf{1839.6} & \hfil- & \hfil- \\
 \hline
\end{tabular}
\end{minipage}
\end{table*}

\section{Proof Outlines}
We briefly sketch the arguments for our stated results.

Theorem~\ref{th:main_result}.\ref{st:W.stability} follows via a similar induction argument as in \cite{gosavi2006boundedness}. On the other hand, Theorem~\ref{th:main_result}.\ref{st:fast.iterate.convergence} follows from the fact that, for a state-action pair $(s, a),$ when $W(s, a)$  is larger (resp. smaller) than $\langle W(s, \cdot), \pi_{\theta_n}(\cdot, s)\rangle,$ then $\theta_{n + 1}(s, a)$ is larger (resp. smaller) than $\theta_n(s, a).$ Note that $\langle W(s, \cdot), \pi_{\theta_n}(\cdot, s)\rangle$ is $W(s, \cdot)$'s average with respect to $\pi_{\theta_n}(\cdot|s).$ Thus, $\pi_{\theta_{n + 1}}$ assigns a relatively higher probability to actions where the $W$ values are larger, from which the desired claim follows. For Theorem~\ref{th:main_result}.\ref{st:slow.iterate.convergence},  we build upon the above results and the ideas in \cite{borkar2009stochastic}, \cite{ramaswamy2016stochastic}, \cite{yaji2020stochastic} and \cite{bhatnagar2023actor}, which discuss the convergence of a stochastic approximation algorithm to a suitably defined limiting DI. Finally, Proposition~\ref{prop:MPE_is_0_of_DI} follows by exploiting the definition of an MPE.

\section{Experiments}
\label{app.sec:Experiments}

We now demonstrate the effectiveness of our algorithm through a numerical example concerning the stochastic inventory control problem. This problem involves managing stock levels---such as cars in a showroom---to meet uncertain daily demand while maximizing overall profits.

For our illustration, we consider an inventory system with a maximum capacity of $M = 2$. We suppose that the procurement (resp. holding\footnote{Holding cost is the daily expense incurred for storing an unsold item.}) cost per item is $c = 500$ (resp. $h = 50$), while the selling price is $p = 900.$ Further, we suppose that the daily demand is a random variable taking values of $0, 1,$ or $2$ with probabilities $0.3,$ $0.2,$ and $0.5,$ respectively. At the start of day $n,$ the inventory manager gets to see the current stock level $s_n$ and then decide on the number of new items $a_n$ to (immediately) procure to meet the (uncertain) demand $d_n$ for that day; the capacity constraint implies that $s_n + a_n$ can be at most $2.$ Hence, the reward obtained for day $n$ equals 
$r_n(s_n, a_n) =900 \times \min\{s_n + a_n, d_n\} - 500 \times a_n - 50 \times \max\{s_n + a_n - d_n, 0\}.$ In turn, starting with an initial stock of $s$ and initial procurement of $a,$ the QH Q-value function of a procurement policy $\pi$ equals the expression on the RHS of \eqref{eq:QH_Q-value_fn}. For our illustration, we suppose $\sigma=0.3$ and $\gamma = 0.9$.


We ran our proposed algorithm multiple times and it identified three different MPEs. These policies are given in Table~\ref{tab:policy_matrices}, and their corresponding QH Q-value functions are given in Table~\ref{tab:IC.Q.values}. As can be seen, MPEs are not unique, and the overall profits associated with different MPEs can vary. For example, $\pi^*_3$ yields the highest returns in our setup. Lastly, note that, at any state, each MPE assigns positive probabilities only to actions with the highest Q-values, although these probabilities need not be equal.

Separately, we also ran the variant of the classical critic-actor method for QH discounting, the one where the actor update is along the gradient of the state value function as described in \eqref{th:PG-QH} (with the initial state distribution being uniform). This yielded the policy $\pi^*,$ given in Table~\ref{tab:policy_matrices}. 
Its QH Q-value function is given in Table~\ref{tab:IC.Q.values}.  

Unlike any of the MPEs, observe that $\pi^*$ does not share a greedy relation with its QH Q-value function. This implies that, e.g., starting from zero stock, the highest overall return is obtained if the stock is raised to $1$ on day $0,$ and to $2$ thereafter. Thus, there is an incentive for deviation, as explained in Section~\ref{sec:introduction}. If the store manager is naive or has self-control issues and deviates from $\pi^*$ every day, then the resulting policy would be $\pi_\text{n}$ (see Table \ref{tab:policy_matrices}). From its Q-values in Table~\ref{tab:IC.Q.values}, it can be seen that $\pi_\text{n}$ yields the lowest overall returns, compared to $\pi^*$ and, importantly, to any of the MPEs. We emphasize that if the manager and their future selves agree to follow an MPE, the current self would have no incentive for deviation. 

\section{Conclusion and Future Directions}
In this work, we have developed the first model-free RL algorithm for finding an MPE in the QH-discounting setting. Our proposed algorithm, based on a two-timescale stochastic approximation, overcomes challenges posed by the lack of a Bellman-type contraction for an MPE's Q-value. We demonstrated its effectiveness in an inventory control problem. Future directions could explore extending this framework to more complex and larger state-action spaces, improving convergence guarantees, and investigating alternative discounting models to capture more nuanced human decision-making behaviors.

\section{Acknowledgements}
We thank Dr. Jayakumar Subramanian (Adobe India), Prof. Mathukumalli Vidyasagar (IIT Hyderabad), and Prof. Shalabh Bhatnagar (IISc Bengaluru) for their insightful discussions and valuable feedback. This significantly contributed to improving the overall quality of this work.

\bibliography{references}

\begin{thebibliography}{10}

\bibitem{ainslie1975specious}
G.~Ainslie.
\newblock Specious reward: a behavioral theory of impulsiveness and impulse control.
\newblock {\em Psychological bulletin}, 82(4):463, 1975.

\bibitem{aubin2012differential}
J.~Aubin and A.~Cellina.
\newblock Differential inclusions: Set-valued maps and viability theory, springer science \& business media.
\newblock {\em Berlin, Germany}, 2012.

\bibitem{Bertsekas}
D.~Bertsekas.
\newblock Reinforcement learning and optimal control.
\newblock {\em Athena Scientific}, 2019.

\bibitem{bhatnagar2023actor}
S.~Bhatnagar, V.~S. Borkar, and S.~Guin.
\newblock Actor-critic or critic-actor? a tale of two time scales.
\newblock {\em IEEE Control Systems Letters}, 2023.

\bibitem{borkar2009stochastic}
V.~S. Borkar.
\newblock {\em Stochastic approximation: a dynamical systems viewpoint}, volume~48.
\newblock Springer, 2009.

\bibitem{cropper1992rates}
M.~L. Cropper, S.~K. Aydede, and P.~R. Portney.
\newblock Rates of time preference for saving lives.
\newblock {\em The American Economic Review}, 82(2):469--472, 1992.

\bibitem{dhami2016foundations}
S.~Dhami.
\newblock {\em The foundations of behavioral economic analysis}.
\newblock Oxford University Press, 2016.

\bibitem{frederick2002time}
S.~Frederick, G.~Loewenstein, and T.~O’donoghue.
\newblock Time discounting and time preference: A critical review.
\newblock {\em Journal of economic literature}, 40(2):351--401, 2002.

\bibitem{gopalan2022demystifying}
A.~Gopalan and G.~Thoppe.
\newblock Demystifying approximate value-based rl with $\epsilon$-greedy exploration: A differential inclusion view, 2023.

\bibitem{gosavi2006boundedness}
A.~Gosavi.
\newblock Boundedness of iterates in q-learning.
\newblock {\em Systems \& control letters}, 55(4):347--349, 2006.

\bibitem{Nowak}
A.~Jaśkiewicz and A.~S. Nowak.
\newblock Markov decision processes with quasi- hyperbolic discounting.
\newblock {\em Finance and Stochastics}, 25(2):189--229, 2021.

\bibitem{Laib}
D.~Laibson.
\newblock Golden eggs and hyperbolic discounting.
\newblock {\em The Quarterly Journal of Economics}, 112(2):443--478, 1997.

\bibitem{loewenstein1992anomalies}
G.~Loewenstein and D.~Prelec.
\newblock Anomalies in intertemporal choice: Evidence and an interpretation.
\newblock {\em The Quarterly Journal of Economics}, 107(2):573--597, 1992.

\bibitem{phelps1968second}
E.~Phelps and R.~Pollak.
\newblock On second-best national saving and game-equilibrium growth.
\newblock {\em The Review of Economic Studies}, 35(2):185, 1968.

\bibitem{ramaswamy2016stochastic}
A.~Ramaswamy and S.~Bhatnagar.
\newblock Stochastic recursive inclusion in two timescales with an application to the lagrangian dual problem.
\newblock {\em Stochastics}, 88(8):1173--1187, 2016.

\bibitem{sutton2018reinforcement}
R.~S. Sutton and A.~G. Barto.
\newblock {\em Reinforcement learning: An introduction}.
\newblock MIT press, 2018.

\bibitem{sutton1999policy}
R.~S. Sutton, D.~McAllester, S.~Singh, and Y.~Mansour.
\newblock Policy gradient methods for reinforcement learning with function approximation.
\newblock {\em Advances in neural information processing systems}, 12, 1999.

\bibitem{thaler1981some}
R.~Thaler.
\newblock Some empirical evidence on dynamic inconsistency.
\newblock {\em Economics letters}, 8(3):201--207, 1981.

\bibitem{yaji2020stochastic}
V.~G. Yaji and S.~Bhatnagar.
\newblock Stochastic recursive inclusions in two timescales with nonadditive iterate-dependent markov noise.
\newblock {\em Mathematics of Operations Research}, 45(4):1405--1444, 2020.

\end{thebibliography}
\bibliographystyle{abbrv}
\end{document}